\documentclass[runningheads]{llncs}
\usepackage{graphicx}
\usepackage{times}
\usepackage{latexsym}
\usepackage{cite}
\usepackage{url}
\usepackage{hyperref}

\usepackage{tabularx}
\usepackage[compatibility=false]{caption}
\usepackage{amsmath,xparse,mleftright}
\usepackage{multirow}
\begin{document}
\title{Reinforced Rewards Framework for Text Style Transfer}
%
%\author{Anonymous Author(s)}
\author{Abhilasha Sancheti\inst{1}\thanks{This work was done while the authors were working at Adobe Research, Bangalore, India.} \and
Kundan Krishna\inst{2}\protect\footnotemark[1] \and
Balaji Vasan Srinivasan\inst{3} \and
Anandhavelu Natarajan\inst{3} }
\authorrunning{Sancheti et al.}
%\institute{Anonymous Affiliation(s)}
\institute{University of Maryland, College Park, USA \\
 \email{sancheti@cs.umd.edu} \and
Language Technologies Institute, Carnegie Mellon University, Pittsburgh, USA \\
\email{kundank@andrew.cmu.edu}
\and
Adobe Research, Bangalore, India\\
\email{{balsrini,anandvn}@adobe.com}}
\maketitle              % typeset the header of the contribution
\begin{abstract}
Style transfer deals with the algorithms to transfer the stylistic properties of a piece of text into that of another while ensuring that the core content is preserved. There has been a lot of interest in the field of text style transfer due to its wide application to tailored text generation. Existing works evaluate the style transfer models based on content preservation and transfer strength. In this work, we propose a reinforcement learning based framework that directly rewards the framework on these target metrics yielding a better transfer of the target style. We show the improved performance of our proposed framework based on automatic and human evaluation on three independent tasks: wherein we transfer the style of text from formal to informal, high excitement to low excitement, modern English to Shakespearean English, and vice-versa in all the three cases. Improved performance of the proposed framework over existing state-of-the-art frameworks indicates the viability of the approach. 

\keywords{Style Transfer  \and Rewards \and Content Preservation \and Transfer Strength}
\end{abstract}
\section{Introduction} \label{sec: introduction}
Text style transfer deals with transforming a given piece of text in such a way that the stylistic properties change to that of the target text while preserving the core content  of the  given text. This is an active area of research because of its wide applicability in the field of content creation including news rewriting, generating messages with a particular style to maintain the personality of a brand, etc. The stylistic properties may denote various linguistic phenomenon, from syntactic changes \cite{W17-4902,xu2012paraphrasing} to sentiment modifications \cite{shen2017style,li2018delete,fu2017style} or extent of formality in a sentence \cite{rao2018dear}.

Most of the existing works in this area either use copy-enriched sequence-to-sequence models \cite{W17-4902} or employ an adversarial \cite{shen2017style,fu2017style,P18-1080} or much simpler generative approaches \cite{li2018delete} based on the disentanglement of style and content in text. On the other hand, more recent works like \cite{subramanian2018multiple} and \cite{dai2019style} perform the task of style transfer without disentangling style and content, as practically this condition cannot always be met. However, all of these works use word-level objective function (eg. cross-entropy) while training which is inconsistent with the desired metrics (content preservation and transfer strength) to be optimized in style transfer tasks. These metrics are generally calculated at a
sentence-level and use of word level objective functions is not sufficient. Moreover, discreteness of these metrics makes it even harder to directly optimize the model over these metrics.

Recent advancements in Reinforcement Learning and its effectiveness in various NLP tasks like sequence modelling \cite{keneshloo2018deep}, abstractive summarization \cite{paulus2017deep}, and a related one machine translation \cite{D18-1397} have motivated us to leverage reinforcement learning approaches in style transfer tasks. 

In this paper, we propose a reinforcement learning (RL) based framework which adopts to optimize sequence-level objectives to perform text
style transfer. Our reinforced rewards framework is based on a sequence-to-sequence model with attention \cite{bahdanau2014neural,luong2015effective} and copy-mechanism \cite{W17-4902} to perform the task of text style transfer. The sentence generated by this model along with the ground truth sentence is passed to a content module and a style classifier which calculates the metric scores to finally obtain the reward values. These rewards are then propagated back to the sequence-to-sequence model in the form of loss terms.

The rest of our paper is organized as follows:
we discuss related work on text style transfer in Section \ref{sec: related}. The proposed reinforced rewards framework is introduced in Section \ref{sec: model}. We evaluate our framework and report the results on formality transfer task in Section \ref{sec: experiments-formality}, on affective dimension like excitement in Section \ref{sec: experiments-beyond} and on Shakespearean-Modern English corpus in Section \ref{sec: experiments-beyond-affect}. In Section \ref{sec: discussion}, we discuss few qualitative sample outputs. Finally, we conclude the paper in Section \ref{sec: conclusion}.
\section{Related Work} \label{sec: related}
Style transfer approaches can be broadly categorized as style transfer with parallel corpus and style transfer with non-parallel corpus.

Parallel corpus consists of input-output sentence pairs with mapping. Since such corpora are not readily available and difficult to curate, efforts here are limited. \cite{xu2012paraphrasing} introduced a parallel corpus of $30$K sentence pairs to transfer Shakespearean English to modern English and benchmark various phrase-based machine translation methods for this task. \cite{W17-4902} use a copy-enriched sequence-to-sequence approach for Shakespearizing  modern English and show that it outperforms the previous benchmarks by \cite{xu2012paraphrasing}. Recently, \cite{rao2018dear} introduced a parallel corpus of formal and informal sentences and benchmark various neural frameworks to transfer sentences across different formality levels.  
Our approach contributes in this field of parallel style transfer and extends the work by \cite{W17-4902} by directly optimizing the metrics used for evaluating the style transfer tasks.

Another class of explorations are in the area of non-parallel text style transfer \cite{shen2017style,fu2017style,li2018delete,P18-1080} which does not require mapping between the input and output sentences. \cite{fu2017style} compose a non-parallel dataset for paper-news titles and propose models to learn separate representations for style and content using adversarial frameworks. \cite{shen2017style} assume a shared latent content distribution across a given corpora and propose a method that leverages refined alignment of latent representations to perform style transfer. \cite{li2018delete} define style in terms of attributes (such as, sentiment) localized to parts of the sentence and learn to disentangle style from content in an unsupervised setting. Although these approaches perform well on the transfer task, content preservation is generally observed to be low due to the non-parallel nature of the data. Along this line, parallel style transfer approaches have shown better performance in benchmarks despite the data curation challenges \cite{rao2018dear}. 

Style transfer models are primarily evaluated on \textbf{content preservation} and \textbf{transfer strength}. But the existing approaches do not optimize on these metrics and rather teach the model to generate sentences to match the ground truth. This is partly because of the reliance on a differentiable training objective and discreteness of these metrics makes it challenging to differentiate the objective. Leveraging recent advancements in reinforcement learning approaches, we propose a reinforcement learning based text style transfer framework which directly optimizes the model on the desired evaluation metrics. Though there exists some prior work on reinforcement learning for machine translation \cite{D18-1397}, sequence modelling \cite{keneshloo2018deep} and abstractive summarization \cite{paulus2017deep} dealing model optimization for qualitative metrics like Rouge \cite{lin2004rouge}, they do not consider style aspects which is one of the main requirements of style transfer tasks. More recently, efforts \cite{xu2018unpaired,gong2019reinforcement} have been made to incorporate RL in style transfer tasks in a non-parallel setup. However, our work is in the field of parallel text style transfer which is not much explored.

Our work is different from these related works in the sense that we take care of content preservation and transfer strength with the use of a content module (to ensure content preservation) and cooperative style discriminator (style classifier) without explicitly separating content and style. We illustrate the improvement in the performance of the framework on the task of transferring text between different levels of formality \cite{rao2018dear}. Furthermore, we present the generalizability of the proposed approach by evaluating it on a self-curated excitement corpus as well as modern English to Shakespearean corpus \cite{W17-4902}.
\section{Reinforced Rewards Framework} \label{sec: model}
The proposed approach takes an input sentence $x= x_{1}\ldots x_{l}$ from source style $s_{1}$ and translates it to sentence $y= y_{1}\ldots y_{m}$ with style $s_{2}$, where \textit{x} and \textit{y} are represented as a sequence of words. If $x$ is given by ($c_{1},s_{1}$) where $c_{1}$ represents the content and $s_{1}$ the style of the source, our objective is to generate $y=(c_{1},s_{2})$ which has same content as the source but with the target style.

Our approach is based on a copy-enriched sequence-to-sequence framework \cite{W17-4902} which allows the model to retain factual parts of the text while changing the style specific text using an attention mechanism.  
At the time of training, the framework takes in the source style and the target style sentence as input to the attention based sequence-to-sequence encoder-decoder model. The words in the input sentence are mapped into an embedding space and the sentence is encoded into a latent space by the LSTM encoder. The network learns to pay attention to the words in the source sentence and creates a context vector based on the attention. The decoder model is a mixture of RNN and pointer (PTR) network where the RNN predicts the probability distribution over the vocabulary and the pointer network predicts the probability over the words in the input sentence based on the context vector. A weighted average of the two probabilities yields the final probability distribution at time step t given by,
$$
\textstyle
P_{t}(w) = \delta P_{t}^{RNN}(w) + (1-\delta) P_{t}^{PTR}(w),
$$
where $\delta$ is computed based on encoder outputs and previous decoder hidden states. The decoder generates the transferred sentence by selecting the most probable word at each time step. This model is trained to minimize cross entropy loss given by
$$
\textstyle
L_{ml} = - \sum_{t=1}^{m}{\log(p(P_{t}(y_{t}^{*})))},
$$                                         
where \textit{m} is the maximum length of the output sentence and $y_{t}^{*}$ is the ground truth word at time \textit{t} in the transferred sentence.  
\begin{figure}
    \centering
    \includegraphics[width=7.5cm, height=2.5cm]{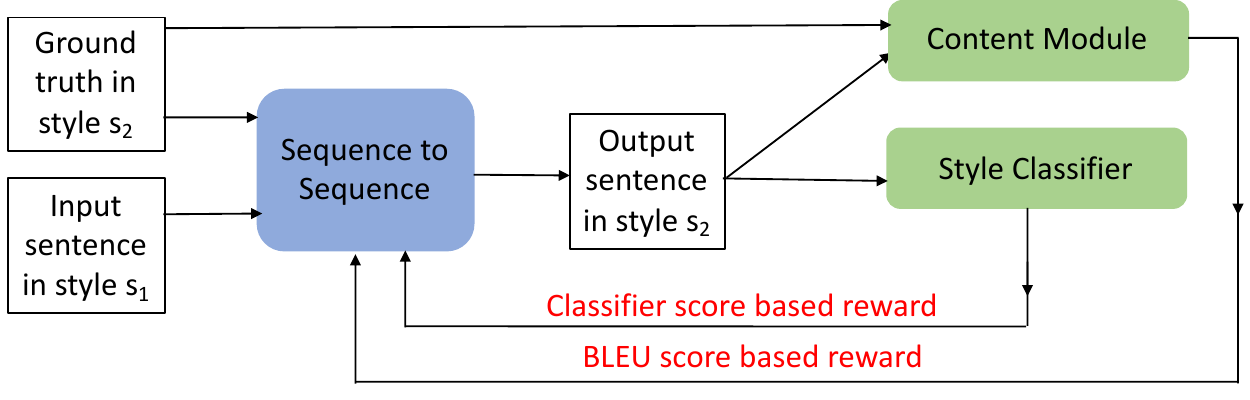}
    \caption{Model overview}
    \label{fig: overview}
\end{figure}
While this framework optimizes for generating sentences close to the ground truth, it does not explicitly teach the network to preserve the content and generate sentences in target style. To achieve this, we introduce a style classifier and a content module which takes in the generated sentence from the sequence-to-sequence model along with the ground truth target sentence to provide reward to the sentence, as shown in Figure \ref{fig: overview}.  We leverage BLEU \cite{papineni2002bleu}  score to measure the reward for preserving content and because of the lack of any formal score for transfer strength, we use a cooperative discriminator to provide score to the generated sentence. This score from the discriminator is used as a measure to reward for transfer strength. These rewards are then back propagated as explicit loss terms to penalize the network for incorrect generation. 
\subsection{Content Module: Rewarding Content Preservation}
To preserve the content while transferring the style, we leverage Self-Critic Sequence Training (SCST) \cite{rennie2017self} approach and optimize the framework with BLEU scores as the reward. SCST is a policy gradient method for reinforcement learning and is used to train end-to-end models directly on non-differentiable metrics. We use BLEU score as reward for content preservation because it measures the overlap between the ground truth and the generated sentences. Teaching the network to favor this would result in high overlap with the ground truth and subsequently preserve the content of the source sentence since ground truth ensures this preservation. 

We produce two output sentences $y^s$ and $y^{'}$, where $y^s$ is sampled from the distribution $p(y_{t}^{s}|y_{1:t-1}^{s},x)$ at each decoding time step and $y^{'}$ (baseline output) is obtained by greedily maximizing the output distribution at each time step. The BLEU score between the sampled and greedy sequences is computed as the reward and the corresponding content-preservation loss is given by, 
$$ 
\textstyle
L_{cp} = (r(y^{'})-r(y^{s}))\sum_{t=1}^{m}{\log(p(y_{t}^{s}| y_{1:t-1}^s,x))},$$                 
where the log term is the log likelihood on sampled sequence and the difference term is the difference between the reward (BLEU score) for the greedily sampled $y^{'}$ and multinomially sampled $y^{s}$ sentences. Note that our formulation is flexible and does not require the metric to be differentiable because rewards are used as weights to the log-likelihood loss. Minimizing $L_{cp}$ is equivalent to encouraging the model to generate sentences which have higher reward as compared to the baseline $y^{'}$ and thus increasing the reward expectation of the model. The framework can now be trained end to end by using this loss function along with the cross entropy loss to preserve the content of the source sentence in the transferred sentence. 
\subsection{Style Classifier: Rewarding Transfer Strength} \label{subsec: ts}
To optimize the model to generate sentences which belong to the target style, it is possible to use a similar loss function as above and use it with the SCST framework \cite{rennie2017self}. However, that will require a formal measure for the target style aspect. Here, we present an alternate framework where such a formal measure is not readily available. We train a convolutional neural network based style classifier as proposed by \cite{kim2014convolutional} on the training dataset. This style classifier predicts the likelihood that an input sentence is in the target style, and the likelihood is taken as a
proxy to the reward for style of a sentence and appended to a discriminator-based loss function extended from \cite{P18-1152}. 
Based on the transfer direction, we add the following term to the cross-entropy loss,
\begin{equation*}
     L_{ts}=
    \begin{cases}
      - \log(1-s(y^{'})),& \text{high to low level} \\
      - \log(s(y^{'})),& \text{low to high level}
    \end{cases}
\end{equation*}
In this formulation, $y^{'}$ is the greedily generated output from the decoder and s($y^{'}$) is the likelihood score predicted by the classifier for $y^{'}$. %We use $y^{'}$ instead of $y^{s}$ because during the inference we output the greedily generated sentence. %\abhilashacomment{addressed} \bvscomment{if possible add a line on why we use $y^{'}$ here and not $y^{s}$} 
When transfer is done from high to low level of style, minimization of $L_{ts}$ will encourage generation of sentences such that the classifier score is as low as possible. When the sentences are transferred from low to high level of style then the formulation ensures that the generated sentences have a score as high as possible. The framework is trained end-to-end using this loss function to generate the sentences which belong to the target style. 
\subsection{Training and Inference}
The overall loss function thus can be written as a combination of the $3$ loss functions, 
%$Loss=  \alpha L_{ml}+ \beta L_{cp}+ \gamma L_{ts}$.
$$ Loss=  \alpha L_{ml}+ \beta L_{cp}+ \gamma L_{ts} $$ 
We train various models using this loss function and different training methodologies (setting $\alpha=1.0$, $\beta=0.125$, $\gamma=1.0$ after hyper-parameter tuning) as described in the next section. During the inference phase, the model predicts a probability distribution over the vocabulary based on the sentence generated so far and the word having the highest probability is chosen as the next word till the maximum length of the output sentence is reached. Note that unlike training phase in which case both the input and ground truth transferred sentences are available to the model, only the input sentence is made available to the model.
%\begin{table*}[bth]
%\centering
%\small
%\begin{tabular}{|c|c|c|c|c|c|c|c|}
%\hline
%& \multicolumn{3}{c|}{\textbf{Exciting to Non-exciting (Informal to Formal)}}&& \multicolumn{3}{c|}{\textbf{Non-exciting to Exciting (Formal to Informal)} } \\
%\hline
%\textbf{Models} & BLEU$\uparrow$  & Avg. Score$\downarrow$ & \% Accuracy$\uparrow$ &&  BLEU$\uparrow$ & Avg. Score$\uparrow$& \% Accuracy$\uparrow$\\
%\hline	
%Transformer &7.70 (12.51) &	0.189 ($0.075^{*}$) & 92.20 ($93.36^{*}$)&&	6.91 (9.93)& 0.541 ($0.889^{*}$)& 60.50 ($89.40^{*}$) \\
%\hline	
%PBMT & 13.52 (26.15) &	0.317 (0.911) &77.07 (8.68)&&	$15.37^{*}$ (27.36)& 0.304 (0.508)& 24.70 (51.09) \\
%\hline	
%CopyNMT  &13.75 (28.04) &	0.206 (0.268) & 91.20 (74.08)&&	7.11 (28.09)& \textbf{0.639} (0.505)& \textbf{82.20} (50.35) \\
%\hline
%CP$\rightarrow$TS &13.46 (22.76)&\textbf{0.179} (\textbf{0.187})&\textbf{94.40} (\textbf{81.71})&&6.88 (25.93)  & \textbf{0.652} (\textbf{0.549}) & \textbf{83.50} (\textbf{54.51})\\
%\hline
%TS$\rightarrow$CP &\textbf{15.33} (\textbf{28.40})&\textbf{0.196} (\textbf{0.270})&\textbf{92.20} (\textbf{74.95})& &\textbf{8.83} (\textbf{29.80}) & 0.597 (\textbf{0.522})& 	74.40 (\textbf{51.69})\\
%\hline
%\end{tabular}\vs\vs\vs
%\caption{Model results for exciting to non-exciting and non-exciting to exciting transfer task with informal to formal and vice-versa in parenthesis. Higher average score implies more excitement(informality).}\label{tab: result-table}\vs\vs\vs\vs\vs\vs
%\end{table*}
\section{Experiments: Reinforcing Formality (GYAFC Dataset)}  \label{sec: experiments-formality}
We evaluate the proposed approach on the GYAFC \cite{rao2018dear} dataset which is a parallel corpus for formal-informal text. We present the transfer task results in both the directions - formal to informal and vice-versa. This dataset (from Entertainment
and Music domain) consists of $\sim$56K informal-formal sentence pairs: $\sim$52K in train, $\sim$1.5K in test and $\sim$2.5K in validation split.

We use both human and automatic evaluation measures for content preservation and transfer strength to illustrate the performance of the proposed approach.

\textbf{Content preservation} measures the degree to which the target style model outputs have the same meaning as the input style sentence. Following \cite{rao2018dear}, we measure preservation of content using BLEU \cite{papineni2002bleu} score between the ground truth and the generated sentence since the ground truth ensures that content of the source style sentence is preserved in it. For human evaluation, we presented $50$ randomly selected model outputs to the Mechanical turk annotators and requested them to rate the outputs on a Likert \cite{bertram2007likert} scale of 6 as  described  in \cite{rao2018dear}.

\textbf{Transfer strength} measures the degree to which style transfer was carried out. We reuse the classifiers that we built to provide rewards to the generated sentences (Section \ref{subsec: ts}). A score above $0.5$ from the classifier represents that the generated sentence belongs to the target style and to the source style otherwise. We define accuracy as the fraction of generated sentences which are classified to be in the target style. The higher the accuracy, higher is the transfer strength. For human evaluation, we ask the Mechanical turk annotators to rate the generated sentence on a Likert scale of $5$ as described in \cite{rao2018dear}.

Following \cite{fu2017style} who illustrate the trade-off between the two metrics - content preservation and transfer strength, we combine the two evaluation measures and present an \textbf{overall score} for the transfer task since both the measures are central to different aspects of text style transfer task. The trade-off arises because the best content preservation can be achieved by simply copying the source sentence. However, the transfer strength in such scenario will be the worst. We compute overall score in the following way 
$$
      \rm{Overall} = \frac{\rm{BLEU} \times \rm{Accuracy}}{\rm{BLEU} + \rm{Accuracy}}
$$
which is similar to F1-score since content preservation can be considered as measuring recall of the amount of source content retained in the target style sentence and transfer strength acts as a measure of precision with which the transfer task is carried out. In the above formulation, both BLEU and accuracy scores are normalized to be between $0$ and $1$.

We first ran an \textbf{ablation study} to demonstrate the improvement in performance of the model with introduction of the two loss terms in the various
settings differing in the way training is being carried out. Below we provide details about each of the settings. \\
\newline\noindent \textbf{CopyNMT: }Trained with $L_{ml}$ 
\newline\noindent \textbf{TS: }Trained with $L_{ml}$ followed by $\alpha L_{ml}+ \gamma L_{ts}$
\newline\noindent \textbf{CP: }Trained with $L_{ml}$ followed by $\alpha L_{ml}+ \beta L_{cp}$
\newline\noindent \textbf{TS+CP: }Trained with $L_{ml}$ followed by $\alpha L_{ml}+\beta L_{cp}+\gamma L_{ts}$ 
\newline\noindent \textbf{TS}$\rightarrow$\textbf{CP: }Trained with $L_{ml}$ followed by $\alpha L_{ml}+ \gamma L_{ts}$ and finally with $\alpha L_{ml}+ \beta L_{cp}$ 
\newline\noindent \textbf{CP}$\rightarrow$\textbf{TS: }Trained with $L_{ml}$ followed by $\alpha L_{ml}+ \beta L_{cp}$ and finally with $\alpha L_{ml}+ \gamma L_{ts}$ \\ \\
Training with $L_{ml}$ alone in all the above settings is done for $10$ epochs with all the hyper-parameters set as default in the off-the-shelf implementation of \cite{W17-4902}. Each of the iterative model training is done using the model with the best performance on validation set for $5$ more epochs. 
\begin{table}[bth]
\centering
%\scriptsize
\begin{tabular}{|c|c|c|c||c|c|c|}
\hline
& \multicolumn{3}{c||}{\textbf{Informal to Formal }}& \multicolumn{3}{c|}{\textbf{Formal to Informal}}\\
\hline
\textbf{Models} & BLEU$\uparrow$  & Accuracy$\uparrow$  & Overall$\uparrow$ &BLEU$\uparrow$  & Accuracy$\uparrow$  & Overall$\uparrow$\\
\hline	
CopyNMT  &  0.263  & 0.774   & 0.196 & 0.280 & 0.503  & 0.180\\
\hline
TS & 0.240 &0.801  & 0.184 & 0.271&0.527  & 0.179\\
\hline
CP & 0.272 &0.749 & 0.199 & 0.281&0.487& 0.178\\
\hline
TS+CP & 0.259&0.772 & 0.194 & 0.271&0.527& 0.179\\
\hline
CP$\rightarrow$TS & 0.227& \textbf{0.817}  & 0.178 &0.259& \textbf{0.5441} & 0.175\\
\hline
TS$\rightarrow$CP &\textbf{0.286} &0.723  & \textbf{0.205} & \textbf{0.298}&0.516 & \textbf{0.189}\\
\hline
\end{tabular}
\vspace{5pt}
\caption{Ablation study to demonstrate the improvement of the addition of the loss terms on formality transfer task. }\label{tab: ablation-table}
\end{table}
 We can observe from Table \ref{tab: ablation-table} that $L_{ts}$ and $L_{cp}$ helps in improving the accuracy which measures transfer strength (TS) and BLEU score which measures content preservation (CP) respectively as compared to CopyNMT. When all the three loss terms are used simultaneously (TS+CP) the resulting performance lies between TS and CP, indicating that there is a trade-off between the two metrics and improvement in one metric is at the cost of another as observed by \cite{fu2017style}. This phenomenon is evident from the results of TS$\rightarrow$CP and CP$\rightarrow$TS where the network gets a bit biased towards the latter optimization. Moreover, improvement in CP$\rightarrow$TS and TS$\rightarrow$CP as compared to TS and CP respectively suggests that incremental training better helps in teaching the framework. Since the performance on both transfer strength and content preservation metrics plays an important role in text style transfer task, we chose TS$\rightarrow$CP, which has the maximum overall score, over the other models for further analysis.
 
 \textbf{Baselines: } We compare the proposed approach TS$\rightarrow$CP against the state-of-the-art cross-aligned autoencoder style transfer approach (Cross-Aligned) by \cite{shen2017style}\footnote{We use the off-the-shelf implementation provided by the authors at \\  \url{https://github.com/shentianxiao/language-style-transfer}}, parallel style transfer approach (CopyNMT) by \cite{W17-4902}\footnote{\url{https://github.com/harsh19/Shakespearizing-Modern-English}} and neural encoder-decoder based transformer model \cite{vaswani2017attention}\footnote{\url{https://github.com/pytorch/fairseq} We also tried using the model proposed by \cite{gong2019reinforcement} to compare against out proposed approach but we couldn't get stable performance on our datasets.} 
 \begin{table}[bth]
\centering
%\scriptsize
\begin{tabular}{|c|c|c|c||c|c|c|}
\hline
& \multicolumn{3}{c||}{\textbf{Informal to Formal }}& \multicolumn{3}{c|}{\textbf{Formal to Informal}}\\
\hline
\textbf{Models} & \textbf{BLEU$\uparrow$}  &  \textbf{Accuracy$\uparrow$}& \textbf{Overall$\uparrow$}  & \textbf{BLEU$\uparrow$}  &  \textbf{Accuracy$\uparrow$}& \textbf{Overall$\uparrow$} \\
%\hline
%Unsupervised-RL \cite{gong2019reinforcement} &0.012 &0.063&0.010&0.008&0.999& 0.007\\
\hline
Transformer \cite{vaswani2017attention} &0.125& \textbf{0.933} & 0.110&0.099 &  \textbf{0.894}& 0.089 \\
\hline
Cross-Aligned \cite{shen2017style} &0.116 &0.670&0.098&0.117&0.766&0.101\\
\hline	
CopyNMT \cite{W17-4902}  &0.263  &  0.774 &0.196 &0.280 &  0.503&0.180  \\
\hline
TS$\rightarrow$CP (Proposed)& \textbf{0.286}&0.723 & \textbf{0.205} & \textbf{0.298}&0.516 & \textbf{0.189}\\
\hline
& \multicolumn{3}{c||}{\textbf{Exciting to Non-exciting}}&\multicolumn{3}{c|}{\textbf{ Non-exciting to Exciting}}\\
%\hline
%Unsupervised-RL \cite{gong2019reinforcement} &0.000 & 0.818 & 0.000 & 0.000& 0.547 &0.000\\
\hline
Transformer \cite{vaswani2017attention} &0.077 & \textbf{0.922} & 0.071 & 0.069& 0.605 &0.062\\
\hline
Cross-Aligned \cite{shen2017style} &0.059 &0.818&0.055&0.061&0.547&0.054\\
\hline	
CopyNMT \cite{W17-4902} &0.143 & 0.919 & 0.124 &0.071& \textbf{0.813} & 0.065 \\
\hline
TS$\rightarrow$CP (Proposed)& \textbf{0.153}  & \textbf{0.922}&  \textbf{0.131} &  \textbf{0.088}  & 0.744 &   \textbf{0.078}\\
\hline
& \multicolumn{3}{c||}{\textbf{Modern to Shakespearean }}&\multicolumn{3}{c|}{\textbf{Shakespearean to Modern}}\\
\hline
%Unsupervised-RL \cite{gong2019reinforcement} &0.000 & 0.283 & 0.000 & 0.000& 0.828 &0.000\\
%\hline
Transformer \cite{vaswani2017attention} &0.027 & \textbf{0.736} & 0.026  &0.046 &  \textbf{0.915}& 0.043  \\
\hline
Cross-Aligned \cite{shen2017style} &0.044 &0.614&0.041&0.049&0.537&0.044\\
\hline
CopyNMT \cite{W17-4902} &0.104  & 0.495 & 0.085 &0.111 & 0.596& 0.093  \\
\hline
TS$\rightarrow$CP (Proposed) & \textbf{0.127} &0.489  & \textbf{0.100} &  \textbf{0.137}&0.567 & \textbf{0.110}\\
\hline
\end{tabular}
\vspace{5pt}
\caption{Comparison of TS$\rightarrow$CP with the baselines on the three transfer tasks in both the directions. All the scores are normalized to be between 0 and 1.} \label{tab: result-table}
\end{table} 

\textbf{Results: }It can be seen from Table \ref{tab: result-table} that even though the transformer model has the best accuracy, it fails in preserving the content. Closer look at the outputs (formal to informal transfer task in Table \ref{tab: sampleformality}) reveal that it generates sentences in target style but the sentences do not preserve the meaning of the input and sometimes are out of context (discussed in the Section \ref{sec: discussion}). Cross-Aligned performs the worst in informal to formal transfer task among all the other approaches because it is generating a lot of unknowns and is not able to preserve content. TS$\rightarrow$CP, on the other hand, has the highest overall score and performs the best in preserving the content. We also observed that the dataset had many sentences containing proper nouns like name of the songs, person or artists. In such cases, copy mechanism helps in retaining the proper nouns whereas other models are not able to do so. This is evident from the higher BLEU scores for our proposed model.
 \begin{table}[!ht]
\centering
%\scriptsize
\begin{tabular}{|c|p{0.9cm}|p{0.9cm}|p{0.9cm}||p{0.9cm}|p{0.9cm}|p{0.9cm}|}
\hline
\textbf{Task} & \multicolumn{3}{c||}{\textbf{ Transfer Strength }} & \multicolumn{3}{c|}{\textbf{Content Preservation}}\\
\hline
&R$>$C&R$>$T&R$>$S&R$>$C&R$>$T&R$>$S\\
\hline
I-F & 88.67 &81.34&70.00& 70.00 &72.67&83.67\\
\hline
F-I & 73.34 &88.67&61.22& 59.34 &79.34&91.80 \\
\hline
E-NE & 64.00 &79.34&68.00& 60.67 &71.34&71.73\\
\hline	
NE-E & 76.67 &70.67&68.00& 69.34 &74.00&70.00\\
\hline
\end{tabular}
\vspace{5pt}
\caption{Human evaluation results of 50 randomly selected model outputs. The values represent the \% of times annotators rated model outputs from TS$\rightarrow$CP (R) as better than the baseline CopyNMT (C), Transformer (T) and Cross-Aligned (S) over the metrics. I-F (E-NE) refers to informal to formal (exciting to non-exciting) task.}\label{tab: human-evaluation}
\end{table} 
 Table \ref{tab: human-evaluation} presents the human evaluation results aggregated over three annotators per sample. It can be seen that in at least 70\% of the cases, annotators rated model outputs from TS$\rightarrow$CP as better than the three baselines on both the evaluated metrics except for the content preservation as compared to CopyNMT in formal to informal task wherein, both the models perform equally good. One reason behind this is that both the models use copy-mechanism.
 %This justifies our selection of TS$\rightarrow$CP because as per humans transfer strength is at least $80\%$ of the times better than CopyNMT and the fall in BLEU score (Table \ref{tab: result-table}) was because of the lesser agreement (word overlap) of model outputs  with the ground truth than the content preservation. 
\section{Experiments: Beyond Formality (Excitement Dataset)}  \label{sec: experiments-beyond}
In order to demonstrate the generalizability of our approach on an affective style dimension like excitement (the feeling of enthusiasm and
eagerness), we curated our own dataset using reviews from Yelp dataset\footnote{https://www.yelp.com/dataset} which is a subset of Yelp's businesses, reviews, and user data. We request human annotators to provide rewrites for given exciting sentences such that they sound as non-exciting/boring as possible. Reviews with rating greater than or equal to $3$ were filtered out and considered as exciting to get the non-exciting/boring rewrites. 
We also asked the annotators to rate the given and transferred sentences on a Likert scale of $1$ (No Excitement at all) to $5$ (Very high Excitement). The dataset thus curated was split into train ($\sim$36K), test (1K) and validation (2K) sets. We evaluate the transfer quality on content preservation and transfer strength metrics as defined in Section \ref{sec: experiments-formality}. 
 
For measuring the transfer strength we train a classifier as described in Section \ref{subsec: ts}.
We use the annotations provided by the human annotators on these sentences to get the labels for the two styles. Sentences with a rating greater than or equal to $3$ were considered as exciting and non-exciting otherwise. 
 
\textbf{Results: }The transfer task in this case is to convert the input sentence with high excitement (exciting) to a sentence with low excitement (non-exciting) and vice-versa. We can observe from Table \ref{tab: result-table} that model performance in the case of excitement transfer task is similar to what we observed in the formality transfer task. However, CopyNMT performs the best in transferring style in case of non-exciting to exciting transfer task because the model has picked up on expressive words (`awesome', `great', and 'amazing') which helps in boosting the transfer strength. 
TS$\rightarrow$CP (with highest overall score) consistently outperforms Cross-Aligned in all the metrics and both the directions. 
Table \ref{tab: human-evaluation} presents the human evaluation results on this transfer task. We notice that 
humans preferred outputs from our proposed model at least 60\% of the times on both the measures as compared to the other three baselines. This provides an evidence that the proposed RL-based framework indeed helps in improving generation of more content preserving sentences which align with the target style.

\section{Experiments: Beyond Affective Elements (English Dataset)} \label{sec: experiments-beyond-affect}
Besides affective style dimensions, our approach can also be extended to other style transfer tasks like converting modern English to Shakespearean English. To illustrate the performance of our model on this task we experimented with the corpus used in \cite{W17-4902}. The dataset consists of $\sim$21K modern-Shakespearean English sentence pairs with $\sim$18K in train, $\sim$1.5K in test and $\sim$1.2K in validation split. We use the same evaluation measures as in the previous two tasks for illustrating the model performance and generalizability of the approach. For this task we present only the automatic evaluation results because manual evaluation of this task is not easy since it requires an understanding of Shakespearean English and finding such population is a difficult task due to limited availability. %\anandcomment{Why ?}
 
\textbf{Results: }We can observe from Table \ref{tab: result-table} that model performance in the case of this transfer task is also similar to what we have observed in the earlier two transfer tasks. Although Cross-Aligned has better accuracy than TS$\rightarrow$CP, it fails to preserve the content (sample 3 of Table \ref{tab: sampleenglish}). Similar is the case with transformer which outperforms others in accuracy but is not able to retain the content (sample 1 of Table \ref{tab: sampleenglish}). TS$\rightarrow$CP outperforms the three baselines in preserving the content with the highest overall score. This establishes the viability of our approach to various types of text style transfer tasks.
 
These experiments further indicate that our proposed reinforcement learning framework improves the transfer strength and content preservation of parallel style transfer frameworks and is also generalizable across various stylistic expression.
\begin{table*}[!hbt]
\centering
%\begin{subfigure}[l]{\linewidth}
\begin{center}
%\scriptsize
\resizebox{\linewidth}{!}{%
\begin{tabular}{|c|l p{7cm} p{9cm}|  } 
\hline
&\textbf{Model} & \textbf{Informal to Formal}& \textbf{Formal to Informal}\\
\hline
1&\textbf{Input} & I want to be on TV! &I do not understand what that has to do with who 's better looking?\\
\hline
&\textbf{Reference}  & I would like to be on television. &I don't know what the hell that has to do with who's better looking but OKAY!\\
&\textbf{Transformer}& I want to be on TV. &i don't know what that 's better looking with the band that do u?\\
&\textbf{Cross-Aligned}&I want to be on TV!&I do n't know that that do to have to talk of more better?\\
&\textbf{CopyNMT}& I would like to be on TV. &I don't understand what that has to do with who 's better looking for?\\
&\textbf{TS$\rightarrow$CP} & I would like to be on TV.&i don't understand what that has to do with who better?\\
\hline
2&\textbf{Input}& when you find out please let me know.&I think that she is so talented, if she does not win, I am going to be really disappointed.\\
\hline
&\textbf{Reference}  & Please let me know when you find out.&he is so talented, if she didn't win, I'd be really disappointed!\\
&\textbf{Transformer}& Keep me informed as soon as you know anything. &I don't think she 's hot, but i'm going to win so she'll win.\\
&\textbf{Cross-Aligned}&If you find out please let me know.&I think she is so funny, she doesn't win, I 'm not sure to be gonna be cute\\
&\textbf{CopyNMT} & When you find out please please please me know?&i think she 's so talented, she 's not that i'm going to be really disappointed.\\
&\textbf{TS$\rightarrow$CP} & Please inform me if you find out.&i think she is so talented , if she doesn't win , I'm gonna be really disappointed\\
\hline
3&\textbf{Input} & I dono I think that is the DUMBEST show EVER!!!!!!& our mother is so unintelligent that she was hit \& by a cop and told the police that she was mugged.\\
\hline
&\textbf{Reference}  & I don't think it 's a very intelligent show.& Your mama is so stupid, she got hit by a cop and told the police that she got mugged.\\
&\textbf{Transformer}& I do not think that the show is appropriate. &your mama is so stupid that she sat on the ocean and said she was a bus. \\
&\textbf{Cross-Aligned}&I think that I am $\langle unk\rangle$ the show $\langle unk\rangle$ $\langle unk\rangle$! &Yo mama is so fat that she had a $\langle unk\rangle$ and got a bunch of that's and she was $\langle unk \rangle$\\
&\textbf{CopyNMT} & I am not sure that is the DUMBEST show EVER!&Your mama is so unintelligent she she hit hit cop and told the police that she was.\\
&\textbf{TS$\rightarrow$CP} & I think that is the DUMBEST show EVER!& your mama is so unintelligent she got hit by a cop and told that she was so .\\
\hline
\end{tabular}
}
\end{center}
%\end{subfigure}
\caption{Sample model outputs and target style reference for  Informal to Formal and  Formal to Informal style transfer task. The first line is the source style sentence (input), second line is the reference output and the following lines correspond to the outputs from the baselines and the RL-based model. \label{tab: sampleformality}}
\end{table*}
\section{Discussion} \label{sec: discussion}
\begin{table*}[!hbt]
\centering
%\begin{subfigure}[l]{\linewidth}
\begin{center}
%\scriptsize
\resizebox{\linewidth}{!}{%
\begin{tabular}{ |c|l l p{8cm}|  }
\hline
&\textbf{Model} & \textbf{Exciting to Non-exciting} & \textbf{Non-exciting to Exciting} \\
\hline
1& \textbf{Input} & delicious food and good environment. &a good choice if you are in the phoenix area.\\
\hline
& \textbf{Reference} & good food and environment. &a must visit if in the phoenix area.\\
&\textbf{Transformer}& i recommend this food. &if you're in the phoenix area, this is the place to go.\\
&\textbf{Cross-Aligned}&good food and good drinks.&a great spot if you 're in the area area.\\
&\textbf{CopyNMT} & the food was good. &this is a great choice of if you are in the phoenix area.\\
&\textbf{TS$\rightarrow$CP} & good food and atmosphere. &if you're in the phoenix area, this is a great choice if you're in the phoenix area .\\
\hline
2& \textbf{Input} & our server alisha was amazing. &the food menu is reasonable and happy hour specials are good.\\
\hline
&\textbf{Reference} & our server alisha did a good job.&reasonable food menu and great happy hour specials.\\
&\textbf{Transformer}& our server was good.& they have a great happy hour menu and the food is very good. \\
&\textbf{Cross-Aligned}&our server server was good&the food is great and happy hour prices are awesome\\
&\textbf{CopyNMT} & our server was good.&the food menu is great and the food is amazing.\\
&\textbf{TS$\rightarrow$CP} & our server alisha was very good.& the food menu is reasonable and happy hour specials are great.\\
\hline
3&\textbf{Input}  & the patio is amazing too. &acceptable food and beers with live music sometimes.\\
\hline
&\textbf{Reference}& i like the patio also.&good food and great beers with occasional live music.\\
&\textbf{Transformer}& the patio ... . great. &live bands, good food and great beer.\\
&\textbf{Cross-Aligned}&the patio is pretty good. &awesome food and great selection of music and music\\
&\textbf{CopyNMT} & the patio is good.&great food and great drinks and live music.\\
&\textbf{TS$\rightarrow$CP} & the patio is good.&great food, great beers, and great music.\\
\hline
\end{tabular}
}
\end{center}
\caption{Sample model outputs and target style reference for Exciting to Non-exciting and Non-exciting to Exciting style transfer task. The first line is the source style sentence (input), second line is the reference output and the following lines correspond to the outputs from the baselines and RL-based model. \label{tab: sample}}
\end{table*}
\begin{table*}[!hbt]
\centering
%\begin{subfigure}[l]{\linewidth}
\begin{center}
\resizebox{\linewidth}{!}{%
\begin{tabular}{|c|l p{6cm} p{9cm}|  } 
\hline
&\textbf{Model} & \textbf{Modern to Shakespearean}& \textbf{Shakespearean to Modern}\\
\hline
1&\textbf{Input} & Don't you see that I'm out of breath? &Good morrow to you both.\\
\hline
&\textbf{Reference}  & Do you not see that I am out of breath? &Good morning to you both.\\
&\textbf{Transformer}& Do you not hear me? &Good morning to you.\\
&\textbf{Cross-Aligned}&Do you not think I had out of breath?&Good morrow to you. \\
&\textbf{CopyNMT} & Do not see see I breath of breath? &Good morning, you both.\\
&\textbf{TS$\rightarrow$CP} & Do you not see that I am out of breath?&Good morning to you both.\\
\hline
2&\textbf{Input}& Do you love me?&Well, well, thou hast a careful father, child.\\
\hline
&\textbf{Reference}  & Dost thou love me?&Well, well, you have a careful father, child.\\
&\textbf{Transformer}& Do you love me?&Well, good luck.\\
&\textbf{Cross-Aligned}&Dost thou love me?&Well, sir, be a man, Give it this.\\
&\textbf{CopyNMT} & Do you love?&Well, well, you hast a father father, child.\\
&\textbf{TS$\rightarrow$CP} & Dost thou love me?&Well, well, you have a careful father, child.\\
\hline
3&\textbf{Input} & Come here, man.& Thou know'st my daughter's of a pretty age.\\
\hline
&\textbf{Reference}  & Come hither, man.& You know how young my daughter is.\\
&\textbf{Transformer}& Come, man. &You are my daughter. \\
&\textbf{Cross-Aligned}&Come hither, Iago.&You know how noble my name is.\\
&\textbf{CopyNMT} & Come hither, man.&You know'st my daughter's age.\\
&\textbf{TS$\rightarrow$CP} & Come hither, man.& You're know'st my daughter's of a pretty age.\\
\hline
\end{tabular}
}
\end{center}
\caption{Sample model outputs and target style reference for  Modern to Shakespearean English and  Shakespearean to Modern English transfer task. The first line is the source style sentence (input), second line is the reference output and the following lines correspond to the outputs from the baselines and the RL-based model. \label{tab: sampleenglish}} \vspace{-4pt}
\end{table*}
In this section, we provide few qualitative samples from the baselines and the proposed reinforcement learning based model.
We can observe from the transformer model output for Input $1$ and $2$ in formal to informal column of Table \ref{tab: sampleformality} that it generates sentences with correct target style but does not preserve the content. It either adds random content or deletes the required content (`band' instead of `better' in $1$ and `hot' instead of `talented' in $2$). As mentioned earlier, in sample output $3$ of Table \ref{tab: sampleformality}, Cross-Aligned is unable to retain the content and tend to generate unknown tokens. CopyNMT, even though is able to preserve content, tend to generate repeated token like 'please' in sample input $2$ (informal to formal task) which results in lower BLEU score than our proposed approach. 
 Transformer model outputs for exciting to non-exciting task in samples $1$ and $2$ of Table \ref{tab: sample}, miss specific content words like `environment' and `alisha' respectively. However, it is able to generate the sentences in target style.  Similary, Cross-Aligned and CopyNMT are also not able to retain the name of the server in sample $2$ of Table \ref{tab: sample}. 
Sample $2$ of Shakespearean to Modern English and $1$ of Modern to Shakespearean English task in Table \ref{tab: sampleenglish} provide evidence for high accuracy and lower BLEU scores for transformer model. 
From sample $2$ of Shakespearean to modern English transfer task, we can observe that Cross-Aligned although can generate the sentence in the target style is not able to preserve the entities like 'father' and 'child'. On the other hand, TS$\rightarrow$CP can not only generate the sentences in the target style but is also able to retain the entities.
There are few cases when CopyNMT is better in preserving the content as compared to other models, for instance, sample $1$ of formal to informal transfer task and sample $3$ of non-exciting to exciting transfer task since it leverages copy-mechanism.

%\anandcomment{Point you are trying to drive here is not clear- Are you saying that the fact that there are lexical changes is something interesting ? - i would say the fact that there are more than lexical changes is what is interesting, isn't it ?}
Another point to notice is the lexical level changes made to reflect the target style. For example, the use of `would', `don't' and `inform' instead of  `want', `dono' and `let me know' respectively for transforming informal sentences into formal ones. Use of colloquial words like `u', `gonna' and `mama' for converting the formal sentences to informal can be observed from the sample outputs. Not only lexical level changes but structural transformations can also be observed as in 'Please inform me if you find out'. In case of excitement transfer task, use of strong expressive words like `amazing' and `great' makes the sentence sound more exciting while less expressive words such as `okay' and `good' makes the sentence less exciting. Use of  `thou' for you and `hither' for here are more frequently used in Shakespearean English than in modern English. These sample outputs indeed provide an evidence that our model is able to learn these lexical or structural level differences in various transfer tasks, be it formality, beyond formality or beyond affective dimensions. 

%We can also notice that TS$\rightarrow$CP is not only able to generate sentences in the desired style but is also able to preserve the content in most of the cases.

\section{Conclusion and Future Work} \label{sec: conclusion}
The primary contribution of this work is a reinforce rewards based sequence-to-sequence model which explicitly optimizes over content preservation and transfer strength metrics for style transfer with parallel corpus. Initial results are promising and generalize to other stylistic characteristics as illustrated in our experimental sections. Leveraging this approach for simultaneously changing multiple stylistic properties (for e.g. high excitement and low formality) is a subject of further research.

\bibliographystyle{splncs04}
\bibliography{mybibliography}

\end{document}